\documentclass{article}

\usepackage{times}
\usepackage{graphicx} % more modern
\usepackage{natbib}

\usepackage{algorithm}
\usepackage{algorithmic}

\usepackage{tikz}
\usepackage{pgfplots}
\usepackage{amsmath}
\usepackage{subcaption}
\usepackage{adjustbox}
\usepackage{multirow}
\usepackage{enumitem}
\usepackage[accepted]{icml2017}

\icmltitlerunning{Generative Knowledge Transfer for Neural Language Models}

\begin{document} 

\twocolumn[
\icmltitle{Generative Knowledge Transfer for Neural Language Models}



% you can specify symbols, otherwise they are numbered in order

\begin{icmlauthorlist}
\icmlauthor{Sungho Shin}{goo}
\icmlauthor{Kyuyeon Hwang}{goo}
\icmlauthor{Wonyong Sung}{goo}

\end{icmlauthorlist}

\icmlaffiliation{goo}{Department of Electrical and Computer Engineering, Seoul National University,  Seoul, 08826 Korea}

\icmlcorrespondingauthor{Sungho Shin}{sungho.develop@gmail.com}
\icmlcorrespondingauthor{Wonyong Sung}{wysung@snu.ac.kr}

\icmlkeywords{Generative knowledge distillation, Privacy, Language model, RNN, LSTM, Text privacy}

\vskip 0.3in
]



\printAffiliationsAndNotice{}  % leave blank if no need to mention equal contribution
%\footnotetext{hi}

\begin{abstract} 
In this paper, we propose a generative knowledge transfer technique that trains an RNN based language model (student network) using text and output probabilities generated from a previously trained RNN (teacher network). The text generation can be conducted by either the teacher or the student network. We can also improve the performance by taking the ensemble of soft labels obtained from multiple teacher networks. This method can be used for privacy conscious language model adaptation because no user data is directly used for training. Especially, when the soft labels of multiple devices are aggregated via a trusted third party, we can expect very strong privacy protection.
\end{abstract} 

\section{Introduction}
\label{sec:introduction}
Neural network based language models (LMs) are used in many fields such as speech recognition, chatbot, sentence completion and machine translation~\cite{mikolov2010recurrent, serban2015building, spithourakis2016numerically, mirowski2015dependency, cho2014learning}. Training such LMs requires a large amount of training data. A straight-forward way of gathering a large amount of training data is to collect user data through mobile or the internet connected devices. However, since device users are increasingly reluctant to leak their privacy, it becomes important to collect data while protecting the privacy. Even after training the LM once, it needs to be updated for the purpose of user adaptation or adding new expression. However, it is not desired to use the user data directly.

Instead of collecting sensitive user data, the model parameters of a neural network adapted to a user can be used for training a new model by the knowledge transfer method~\cite{hinton2015distilling}. However, this approach can also cause an unwanted privacy violation by an adversary through a machine learning model attack. If the adversary can access the machine learning model, the output of the model can be used to restore the face of the individual used in the training~\cite{fredrikson2015model}. In the case of text data, similar attacks are possible because an LM can be used as a text generator for generating the sensitive user data used for training the model~\cite{sutskever2011generating,graves2013generating}. Therefore, even the model parameters trained with sensitive data should not allow direct access to the adversary. 

Ensemble of knowledge aggregation can increase the security of personal data. Ensemble methods combine the results of multiple classifiers to improve the performance of machine learning algorithms~\cite{dietterich2000ensemble}. Several studies trained private classifiers to produce a final distributable classifier in various ways, such as averaging the model parameters of teacher networks~\cite{pathak2010multiparty}, training hard labels by voting the ensemble of all the teachers~\cite{papernot2016semi}, or using soft labels~\cite{hamm2016learning}. In this process, the final classifier mixes random noise, such as Laplacian or Gaussian noise, to hide information about a specific person and achieve strong differential privacy~\cite{dwork2006calibrating, dwork2014algorithmic}.

In this paper, we propose a method that efficiently transfers personal text data information for training a recurrent neural network (RNN) based LM while minimizing privacy infringement. This method sends the soft labels generated by the teacher networks instead of sending personal data or model parameters, and the soft outputs obtained from the individual users are aggregated for training a student network by a reliable third party. The proposed GKT trains the student network using only the generated data and labels without the original training data, by operating RNN LMs as a generative model. The text generation can be conducted by either the teacher or the student network.

This paper is composed as follows. Section~\ref{sec:related} introduces the related work and Section~\ref{sec:generative} describes the proposed GKT. Section~\ref{sec:application} describes the GKT by ensemble of multiple LMs, Section~\ref{sec:experimental} shows the experimental results, and Section~\ref{sec:concluding} concludes this paper.

\begin{figure}[t]
\centering
\begin{subfigure}{\linewidth}
  \centering
 \includegraphics[width=\linewidth]{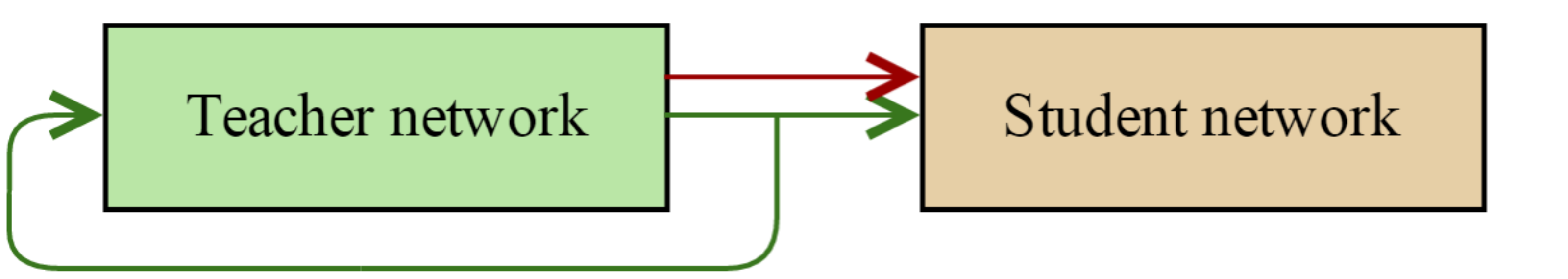}
  \caption{Teacher-driven generative knowledge transfer (TDGKT)}
\label{fig:1_a}
\end{subfigure}%

\begin{subfigure}{\linewidth}
  \centering
  \includegraphics[width=\linewidth]{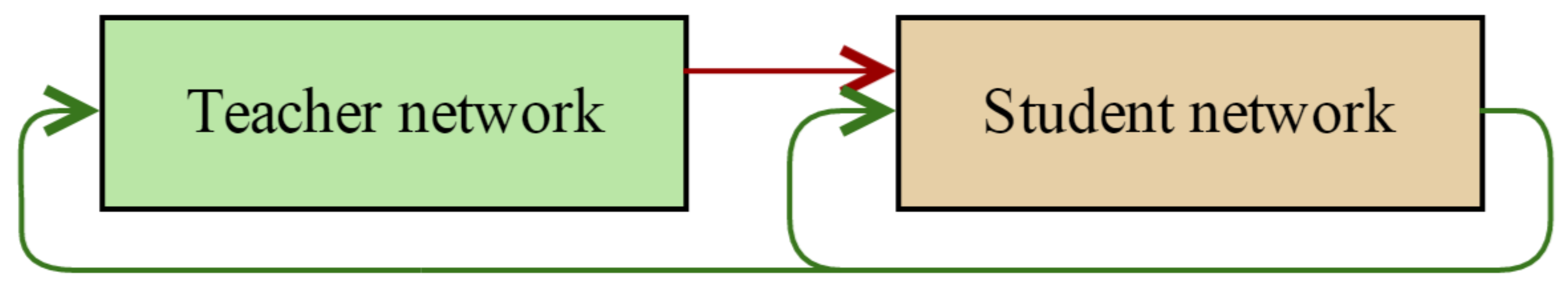}
  \caption{Student-driven generative knowledge transfer (SDGKT)}
\label{fig:1_b}
\end{subfigure}
\caption{Two different schemes of GKT. Text sequence generation (green lines) can be produced by the teacher (TDGKT) or the student network (SDGKT).}
\label{fig:1}
\end{figure}

\section{Related Work on Knowledge Transfer}
\label{sec:related}
Utilizing the knowledge contained in the previously trained networks has been of much interest for the application to network compression or pretraining. In an early work for network compression, a previously trained model is used to label a large unlabeled dataset for producing a much larger training set~\citep{bucilu2006model}. Another related work is a knowledge transfer through the hidden Markov model (HMM)~\citep{pasa2014hmm}. An HMM is trained using the original data, and then the generated sequence from the HMM is used for pretraining of an RNN, which is then fine-tuned using the original data. In the Hinton's knowledge distillation~\citep{hinton2015distilling}, the output probabilities of a well-trained network are used as the soft target for training a small network. In FitNet, a thick-shallow model is transformed to a thin-deep model~\citep{romero2014fitnets}. They employ a guided layer in the student network that can be pretrained from the teacher's corresponding hidden layer, and fine-tuned using knowledge distillation. Also, the model change from a fully connected deep neural network (FCDNN) to an RNN or the opposite direction has been tried~\citep{wang2015recurrent,chan2015transferring}.

The main difference between the previous works and ours is the use of the original training data. The previous works try to improve the performance of training by generating more data using the developed model, but the original data is also used.  Our study conducts training only by using the trained network. Thus, the proposed approach can be considered pure knowledge transfer.

\section{Generative Knowledge Transfer for RNN LMs}
\label{sec:generative}
In this section, we explain the generative knowledge transfer (GKT) for training RNN LMs. We employ teacher-student training scheme. The previously trained network is referred to as the teacher network, and the network that learns from the teacher network is called the student network. The generative knowledge transfer (GKT) proposed in this paper train the student network from a teacher network or networks with only the generated data. For RNN LMs, the GKT can be classified as the teacher-driven GKT (TDGKT) and the student-driven GKT (SDGKT) as shown in~\figurename~\ref{fig:1}. Note that soft labels are always created by the teacher network.

The proposed method consists of three steps; (1) train the teacher network using original training data, (2) generate a text sequence and soft labels to train the student network, (3) train the student network using the generated text data and soft labels. 
%\begin{table}[]
%\centering
%\caption{Two types of GKT for RNN LMs.}
%\label{table:summary}
%\begin{tabular}{ccc}
%\hline\hline
%       & TDGKT   & SDGKT   \\ \hline
%Generate data from   & Teacher & Student \\ \hline
%Generate soft lables from & Teacher & Teacher \\ \hline\hline
%\end{tabular}
%\end{table}

\subsection{Teacher-driven Generative Knowledge Transfer (TDGKT)}
\label{subsec:teacher}

In TDGKT, a student network is trained by creating text using a teacher network as shown in~\figurename~\ref{fig:1_a}. In our study, we use a character-level language model (CLM) which is easy to handle out of vocabulary (OOV) words and has a simpler structure than the word-level language model (WLM). Using a CLM, we can predict the next output probabilities as follows~\cite{sutskever2011generating}:
\begin{align}
	P_{\theta_\mathrm{teacher}}(x^{\mathrm{teacher}}_{t+1} \mid x^{\mathrm{teacher}}_{1:t}) = \boldsymbol{\mathrm{y}}^{\mathrm{teacher}}_{t}, \label{eq:tdgkt}
\end{align}
where $\theta_\mathrm{teacher}$ is the trained teacher model, $t$ is the time step, $x$ is the input data sequences, and $\boldsymbol{\mathrm{y}}_t$ is the model output at time step $t$ (after applying a softmax activation function). The teacher network can generate the text sequence by random sampling from the distribution $P_{\theta_\mathrm{teacher}}(x^{\mathrm{teacher}}_{t+1} \mid x^{\mathrm{teacher}}_{1:t})$.

\begin{figure}[t]
\begin{center}
%\includegraphics{CNN}
%\framebox[5.51in]{\includegraphics{CNN}}
%\fbox{\rule[-.5cm]{0cm}{4cm}\includegraphics{CNN} \rule[-.5cm]{4cm}{0cm}}
\includegraphics[width=0.98\linewidth]{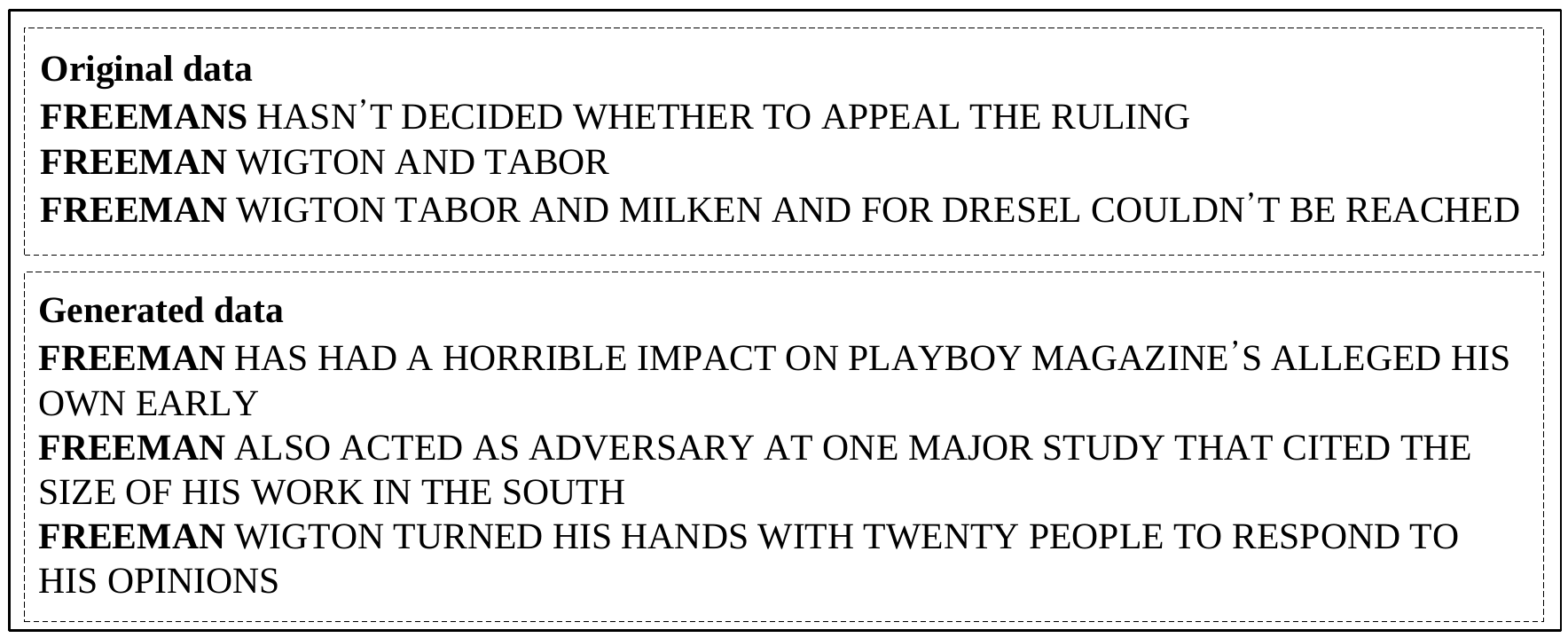}
\end{center}
\caption{Examples of original text and generated text from the teacher network. The first word is a name "FREEMAN", which is included in the original training data.}
\label{fig_2}
\end{figure}
However, since the generated text sequence is quite different from the original text sequence as shown in~\figurename~\ref{fig_2}, the performance of the student network using only the generated text is not as good as that of the original training data. We solve this problem by using softmax output probabilities as soft training labels for the student network. A small example that proves the importance of soft labels in GKT training is given in Section~\ref{subsec:a}.

\subsection{Student-driven Generative Knowledge Transfer (SDGKT)}
\label{subsec:student}
The SDGKT, which is shown in~\figurename~\ref{fig:1_b}, generates the input data sequence from the student network, whereas the soft labels are from the teacher network as in the TDGKT. 
\begin{align}
\begin{split}
	P_{\theta_\mathrm{student}}(x^{\mathrm{student}}_{t+1} \mid x^{\mathrm{student}}_{1:t}) &= \boldsymbol{\mathrm{y}}^\mathrm{student}_{t}\\
	P_{\theta_\mathrm{teacher}}(x^{\mathrm{teacher}}_{t+2} \mid x^{\mathrm{student}}_{1:t+1}) &= \boldsymbol{\mathrm{y}}^\mathrm{teacher}_{t+1} \label{eq:sdgkt}
\end{split}
\end{align}

Note that the text sequence is sequentially sampled from the output of the student network, $P_{\theta_\mathrm{student}}(x^{\mathrm{student}}_{t+1}\mid x^{\mathrm{student}}_{1:t})$. Therefore, the student network needs to be pretrained in order to generate a suitable text sequence unlike TDGKT.

In SDGKT training, there is no concept of `epoch' because the student network can generate data infinitely. Instead, we introduced the concept of `cycle' and `lot'. The student network generates a text sequence with the size of a `lot', feeds it to the teacher network, and trains the student network with the obtained soft labels. A single repetition of this process is called a `single-cycle', and repeating more than two times is called a `multi-cycle'. If the `lot' is 5M characters and the `cycle' is 10, it means that the student network has generated 50M characters and the training has been repeated 10 times each with 5M characters. Note that the sequence is generated by the updated student network, thus the generated sequence will gradually follow the distributions of the original training data learned by the teacher network.

The `multi-cycle' training plays a very important role. For example, the student network can only generate the words in the dataset that have been used for pretraining. Therefore, if the OOV words that are not in the dataset used for pretraining the student network were in the dataset that trained the teacher network, the student network can not learn it properly\footnote{However, student network can learn the OOV words in certain conditions. This situation is shown in Section~\ref{sec:experimental}}. However, if the parameters of the student network are updated multiple times during the training, the student network can progressively generate words from OOV words starting from the first character, eventually learning the whole word corresponding to OOV words.
For example, assume that the word JANUARY is in the data that trained the teacher network, but it is not in the data that pretrained the student network. In this case, the student network gradually generates a portion of the word JANUARY as ``JAN $\rightarrow$ JANU $\rightarrow$ JANUA $\rightarrow$ JANUAR $\rightarrow$ JANUARY'' as the cycle increases during SDGKT training. Thus, we can learn OOV words completely after a few cycles.

\section{Application: Learning Language Models while Preserving User Privacy}
\label{sec:application}
\begin{figure*}[t]
\begin{center}
%\includegraphics{CNN}
%\framebox[5.51in]{\includegraphics{CNN}}
%\fbox{\rule[-.5cm]{0cm}{4cm}\includegraphics{CNN} \rule[-.5cm]{4cm}{0cm}}
\includegraphics[width=0.8\linewidth]{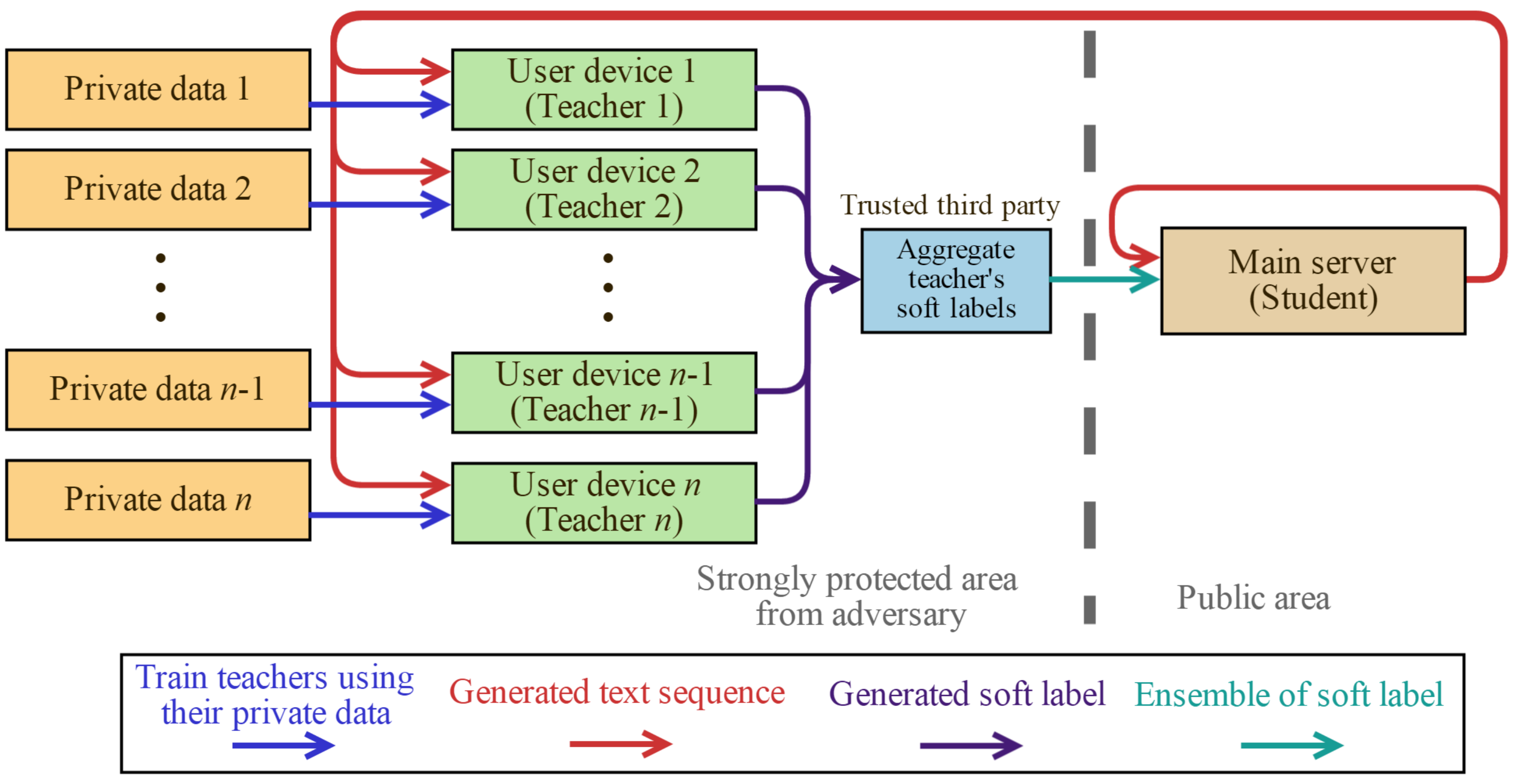}
\end{center}
\caption{Overview of our system for learning the language trends with preserving privacy. (1) The teacher and the student networks are trained using public or in-house data; (2) the teacher networks are fine-tuned (or overfitted) using their private data; (3) the student network generates text sequence and feeds it to the teacher networks to obtain the ensemble of soft labels; (4) the student network is trained using the aggregated soft labels and the generated sequence.}
\label{fig:3}
\end{figure*}

The GKT described in Section~\ref{sec:generative} assumes a single teacher network. In many cases, however, it is necessary to collect data from many users and use them for training. This approach not only helps to improve the performance because it collects data from multiple people, but it also plays a critical role in preserving privacy.
The generative knowledge transfer with multiple teachers proposed in this paper is described in~\figurename~\ref{fig:3}. This figure shows SDGKT-based multiple teacher networks, which can be similarly configured for TDGKT with weaker privacy protection. In this proposed method, the pretrained student network using the public (or in-house) data generates the text sequence and feeds it to the teacher networks for obtaining the soft labels. The soft labels generated by all teacher networks are aggregated together and then applied to the student network. When the soft labels are aggregated using a trusted third party, the privacy can be greatly increased. In this section, the student network is on the main server and the teacher networks are in the user devices.

The proposed method consists of 4 steps as shown in~\figurename~\ref{fig:3}. In the first step, the LM in the main sever (student network) is trained using public data or in-house data. The second step distributes the LM trained in the first step to all the participating user devices. In the third step, each user device fine-tunes (or overfits) its LM with its own private data. The last step is to train the LM on the main server using the ensemble of soft targets obtained from the user devices.
At this time, the LM on the main server generates the text and sends it to the user devices, where the text is applied as the input to the LMs in the devices. In this case, each device generates the soft labels using the given text and sends it to the trusted third party. The trusted third party aggregates the soft labels\footnote{We can also apply differential privacy mechanism~\cite{dwork2014algorithmic} by mixing random noise during the aggregation of soft labels. However, we do not inject random noise because it is out of the scope of this paper.} and sends them to the main server for updating the LM.
 
With this procedure, the main server, which is a service provider, cannot access the original soft labels from each user device, thus it is difficult to infringe on the privacy of the individual users. The trusted third party does not know any information about the input sequence, and thus it cannot violate the privacy of users. In other words, even the trusted third party that collects soft labels does not have the whole information about the individual data. 

Therefore, even if an external hacking attack exposes data from either the trusted third party or the main server, it is difficult for the adversary to utilize the information.  Unless both the main server and the trusted third party are intruded, the privacy of individual users can be kept.

The training method with an ensemble of soft labels was previously proposed for private learning in~\citet{hamm2016learning}. However, GKT training does not require auxiliary unlabeled data, so it has the advantage that no additional human intervention is required. The information gathered from many user devices also allows the server (student network) to adapt the LM for learning new words, expressions, or trendy dialogue styles.

\section{Experimental Results}
\label{sec:experimental}
The RNN-based CLM is used to utilize the OOV words generation capability of the model. The RNN is based on a deep long short-term memory (LSTM) network and $N$x$M$ LSTM means $M$ LSTM layers each containing $N$ memory cells~\cite{hochreiter1997long, graves2013speech}. The input of this CLM is a 30-dimensional vector that is one hot encoded for representing alphabets (A$\sim$Z) and four special symbols ($\textbf{space}$, $\boldsymbol{<eos>}$, $\textbf{'}$, $\textbf{.}$). The output vector represents the probabilities of characters and symbols, and is also represented as a 30-dimensional vector. The CLMs are trained using the Wall Street Journal LM training text with non-verbalized punctuation~\cite{paul1992design}. The RNN training employs the truncated backpropagation through time (BPTT) algorithm~\cite{werbos1990backpropagation} with Adadelta~\cite{zeiler2012adadelta} and Nesterov momentum~\cite{sutskever2013importance}. We report the performance using bits per character (BPC), which is the standard performance measure for CLMs.

\subsection{A Simple TDGKT Example}
\label{subsec:a}
We first examine the effect of the soft labels in GKT training as mentioned in~Section~\ref{subsec:teacher}. \tablename~\ref{table:1} shows the BPCs of teacher and student networks, where the original WSJ text data is used for training. The 256x2 and the 512x2 configurations are chosen as the student networks and 1024x4 as the teacher networks in this experiment.

%\begin{table}[t]
%\centering
%\caption{BPCs of teacher and student networks for the simple TDGKT example. Original WSJ text data is used for training.}
%\label{table:1}
%\begin{tabular}{ccccc}\\
%\hline\hline
% \begin{tabular}[c]{@{}c@{}}             \end{tabular} &\bf{256x2} & \bf{512x2} & \bf{512x4} & \bf{1024x4} \\ \hline
%%\begin{tabular}[c]{@{}c@{}}ACE\\ (Average cross entropy)\end{tabular} & 0.884     & 0.796     & 0.785     & 0.763      \\ \hline
%\begin{tabular}[c]{@{}c@{}}BPC\end{tabular}     & 1.275     & 1.148     & 1.132     & 1.101 \\   \hline\hline
%\end{tabular}
%\end{table}
\begin{table}[t]
\centering
\caption{BPCs of the teacher and student networks for the simple TDGKT example. The original WSJ text data is used for training.}
\label{table:1}
\begin{tabular}{cccc}
\hline\hline
    & \multicolumn{2}{c}{Student} & Teacher \\ \hline
 Size of network  & 256x2         & 512x2        & 1024x4  \\ \hline
BPC & 1.275         & 1.148        & 1.101   \\ \hline\hline
\end{tabular}
\end{table}

\subsubsection{Convergence Speed of Conventional and GKT Training}
\label{subsubsec:trade}
In order to evaluate the performance of the proposed training method, we estimate the difference between the hard and soft targets, and also the effects of using the original training data and the teacher generated sequences. Thus, there are four different experiments. The first one uses the hard target with the original training data. The second one employs the soft target with the original training data. The third method uses the hard target with the teacher generated sequences. The fourth one utilizes the soft target with the teacher generated sequences. The first one represents the conventional training method, while the fourth one is the GKT. 
The experiments are conducted to know the performance of the student network when the amount of the training data generated by the teacher network is limited, such as 10M, 50M,100M, 150M, 215M, and 250M characters. Note that the original data have 215M characters. Although it is possible to generate an infinite length of data using the teacher network, we try to know the efficiency of teacher-student knowledge transfer by limiting the data size. The training is conducted employing many epochs until the performance saturates.

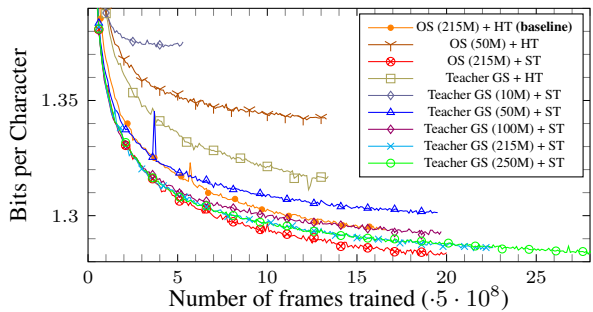
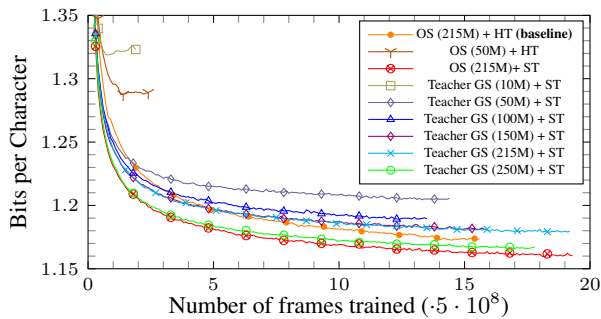
\begin{figure}[t]
\centering
\begin{subfigure}{\linewidth}
\begin{tikzpicture}
    \begin{axis}[
	width=\columnwidth,
	height = 0.6\columnwidth,
	%compat=1.12,
	xmin=0,
	ymin=1.28,
	xmax=28,
	ymax=1.39,
	label style={font=\footnotesize},
	legend style={nodes={scale=0.52, transform shape},at={(0.54,0.98)},anchor=north west},
	tick label style={font=\scriptsize}, 
	domain=1:512, 
	minor x tick num=4, 
	minor y tick num=4, 
	log basis x={10}, 
	xtick pos=both, 
	xtick align=inside, 
	major tick style={line width=0.010cm, black},
	 major tick length=0.10cm,%same as default ]% 
        xlabel=Number of frames trained ($\cdot5\cdot10^8$),
        ylabel=Bits per Character,
	y label style={at={(axis description cs:0.05,0.5)}},
	x label style={at={(axis description cs:0.5,0.1)}}]
	\legend{OS (215M) + HT \textbf{(baseline)}, OS (50M) + HT, OS (215M) + ST, Teacher GS + HT, Teacher GS (10M) + ST, Teacher GS (50M) + ST, Teacher GS (100M) + ST, Teacher GS (215M) + ST, Teacher GS (250M) + ST};
	\addplot[color=orange, mark=*, mark size=1.0pt, solid, mark repeat=15,mark options=solid] file{data/data_256x2_original.txt}; 
	\addplot[color=red!70!green, mark=Mercedes star flipped, mark size=2.5pt, solid, mark repeat=10,mark options=solid] file{data/new_data_256x2_original_50MB.txt}; 
	\addplot[color=red, mark=otimes, mark size=1.6pt, solid, mark repeat=15,mark options=solid] file{data/new_data_original_256x2_KD.txt};%data_256x2_originla_softtarget.txt};
	\addplot[color=yellow!60!black, mark=square, mark size=1.6pt, solid, mark repeat=15,mark options=solid] file{data/data_256x2_mentor_hardtarget.txt}; 
	\addplot[color=blue!60!yellow, mark=diamond, mark size=1.6pt, solid, mark repeat=15,mark options=solid] file{data/data_256x2_mentor_10MB.txt}; 
	\addplot[color=blue, mark=triangle, mark size=1.6pt, solid, mark repeat=15,mark options=solid] file{data/data_256x2_mentor_50MB.txt};
	\addplot[color=red!60!blue, mark=diamond, mark size=1.6pt, solid, mark repeat=15,mark options=solid] file{data/data_256x2_mentor_100MB.txt};
	%\addplot[color=brown, mark=*, mark size=1.6pt, solid, mark repeat=15,mark options=solid] file{data/data_256x2_mentor_150MB.txt};
	\addplot[color=cyan, mark=x, mark size=1.8pt, solid, mark repeat=12,mark options=solid] file{data/new_data_256x2_mentor_210MB.txt};
	\addplot[color=green, mark=o, mark size=1.5pt, solid, mark repeat=15,mark options=solid] file{data/new_data_256x2_mentor_250MB.txt};
    \end{axis}
   \end{tikzpicture}
 \caption{256x2}\label{fig_4_a}
\end{subfigure}
\begin{subfigure}{\linewidth}
\begin{tikzpicture}
    \begin{axis}[
	width=\columnwidth,
	height = 0.6\columnwidth,
	%compat=1.12,
	xmin=0,
	ymin=1.15,
	xmax=20,
	ymax=1.35,
	label style={font=\footnotesize},
	legend style={nodes={scale=0.52, transform shape},at={(0.54,0.98)},anchor=north west},
	tick label style={font=\scriptsize}, 
	domain=1:512, 
	minor x tick num=4, 
	minor y tick num=4, 
	log basis x={10}, 
	xtick pos=both, 
	xtick align=inside, 
	major tick style={line width=0.010cm, black},
	 major tick length=0.10cm,%same as default ]% 
        xlabel=Number of frames trained ($\cdot5\cdot10^8$),
        ylabel=Bits per Character,
	y label style={at={(axis description cs:0.05,0.5)}},
	x label style={at={(axis description cs:0.5,0.1)}}]
	\legend{OS (215M) + HT \textbf{(baseline)}, OS (50M) + HT, OS (215M)+ ST, Teacher GS (10M) + ST, Teacher GS (50M) + ST, Teacher GS (100M) + ST, Teacher GS (150M) + ST, Teacher GS (215M) + ST, Teacher GS (250M) + ST};
	%\legend{OS + HT (baseline), OS + ST,Teacher GS + HT, Teacher GS (10MB) + ST, Teacher GS (50MB) + ST, Teacher GS (100MB) + ST, Teacher GS (150MB) + ST, Teacher GS (210MB) + ST, Teacher GS (250MB) + ST};
	\addplot[color=orange, mark=*, mark size=1.0pt, solid, mark repeat=15,mark options=solid] file{data/data_512x2_original.txt}; 
	\addplot[color=red!70!green, mark=Mercedes star flipped, mark size=2.5pt, solid, mark repeat=10,mark options=solid] file{data/new_data_512x2_original_50MB.txt}; 
	\addplot[color=red, mark=otimes, mark size=1.6pt, solid, mark repeat=15,mark options=solid] file{data/new_data_original_512x2_KD.txt};%data_256x2_originla_softtarget.txt};
	%\addplot[color=green, mark=square, mark size=1.6pt, solid, mark repeat=20,mark options=solid] file{data/data_512x2_mentor_hardtarget.txt}; 
	\addplot[color=yellow!60!black, mark=square, mark size=1.6pt, solid, mark repeat=15,mark options=solid] file{data/data_512x2_mentor_10MB.txt}; 
	\addplot[color=blue!60!yellow, mark=diamond, mark size=1.6pt, solid, mark repeat=15,mark options=solid] file{data/data_512x2_mentor_50MB.txt};
	\addplot[color=blue, mark=triangle, mark size=1.6pt, solid, mark repeat=15,mark options=solid] file{data/data_512x2_mentor_100MB.txt};
	\addplot[color=red!60!blue, mark=diamond, mark size=1.6pt, solid, mark repeat=15,mark options=solid] file{data/data_512x2_mentor_150MB.txt};
	\addplot[color=cyan, mark=x, mark size=1.8pt, solid, mark repeat=12,mark options=solid] file{data/new_data_512x2_mentor_210MB.txt};
	\addplot[color=green, mark=o, mark size=1.5pt, solid, mark repeat=15,mark options=solid] file{data/new_data_512x2_mentor_250MB.txt};
    \end{axis}
   \end{tikzpicture}
 \caption{512x2}\label{fig_4_b}
\end{subfigure}
\caption{Convergence curves in terms of BPC on the validation set in the fixed learning rate of 1e-5. ``OS'' is original training sequence. ``HT'' and ``ST'' are hard and soft targets. ``Teacher G'' is the generated sequences from teacher network and the number in the parenthesis are the sizes of the sequences for performance evaluation.} \label{fig_4}
\end{figure}
\begin{table*}[t]
\centering
\caption{Comparison of TDGKT according to the amount of generated data. The BPCs on the test set are measured.}
\label{table:2}
\begin{tabular}{c|cccccc|cc|c}
\hline\hline
      & \multicolumn{6}{|c|}{TDGKT (GS+ST)} & \multicolumn{2}{c|}{OS+HT (Baseline)} & OS+ST    \\ \hline
Size&10M&50M&100M&150M&215M&250M&50M&215M&215M\\ \hline
256x2 & 1.371       & 1.288       & 1.283       & 1.283   & 1.275 & 1.272 & 1.329 & 1.275 & 1.272       \\ \hline
512x2 & 1.314       & 1.196       & 1.177       & 1.168   & 1.165 & 1.151 & 1.274 & 1.148 & 1.146       \\ \hline\hline
\end{tabular}
\end{table*}
\figurename~\ref{fig_4} shows the training curves of the 256x2 and 512x2 LSTM RNNs for the four different training methods. As for the reference, the training result using the original data, WSJ LM training text, and hard target is given. Also, the training result that uses the generated sequence but not the soft target is also shown. Although the generated data of 900M characters is used for this hard target based teacher-student transfer learning, the result is not sufficient in \figurename~\ref{fig_4_a}. The BPC only reaches to 1.311 with the validation set. The remaining five graphs show the training results using the generated sequence and the soft target of the teacher network, where the data size is intentionally limited. We can find that almost comparable performance can be achieved by applying the proposed method when the generated sequence size is 100 M characters, which is smaller than that of the original data size. The learning speed is even faster than that using the original data. This speed-up is due to the knowledge transfer effect~\citep{hinton2015distilling}. Finally, the \figurename~\ref{fig_4_a} also shows the training result that utilizes both the original data and the soft target, which apparently shows the best results.

In \figurename~\ref{fig_4_b}, the data size is deliberately limited to 10M, 50M, 100M, 150M, 215M, and 250M characters in 512x2 LSTM network which has a higher capacity than the 256x2 LSTM network. The comparable performance can be achieved when the generated sequence size is 215 M characters, but it needs more training frames to reach the baseline performance. With the 250M generated characters, the convergence speed is even faster than that using the original data with the hard targets, and almost reaches the performance of the original data and the soft targets.

The final training results are reported in \tablename~\ref{table:2}. The baseline which is trained using the original sequence and the hard target is 1.329 and 1.275 with 50M and 215M characters, respectively. The proposed method shows better BPC in 50 M characters and achieves the same BPC with 215M characters. Even slightly better BPC is observed in 250M characters, which is the same result with the original sequence and the soft target training. The 512x2 network is also trained fairly well with the TDGKT. However, when we compare the 256x2 and the 512x2 networks, the generative transfer for the latter network seems less satisfactory. The 512x2 network also shows better BPC in 50M characters, but slightly worse result even if the 250M characters are need for the training.

\subsection{Privacy Conscious Language Model Adaptation using Multiple Teacher Devices}
\label{subsec:training}
This sub-section shows the experimental setup and results of the multiple teacher device based training proposed in~Section~\ref{sec:application}.

\subsubsection{Experimental Setup}
\label{subsubsec:experimental}

\textbf{Data partitioning}: The WSJ corpus used in this paper consists of about 215M characters. About 98\% of the data is designated as the training set (210M characters) and the remaining 2\% data is divided into the validation (2M characters) and the test sets (2M characters). Private data for training the user LM was created by intentionally removing some words from the WSJ corpus. To be specific, sentences containing vocabulary for the months (JANUARY, FEBRUARY, ..., and DECEMBER), days of the week (MONDAY, TUESDAY, ..., and SUNDAY), seasons (SPRING, SUMMER, AUTUMN, and WINTER) were separated from the WSJ corpus. The size of the separated private text data is about 29 M characters, which is about 14\% of the original corpus size. As a result, the remaining public (or in-house) data has 181M characters. For performance evaluation, the validation and the test sets are classified into three kinds as follows.
\begin{enumerate}[nolistsep]
\item $\text{Valid}_{\text{private}}$, $\text{Test}_{\text{private}}$: the data consisting of sentences containing the private words in the validation and the test sets, respectively.
\item $\text{Valid}_{\text{public}}$, $\text{Test}_{\text{public}}$: the data consisting of sentences that do not contain the private words in the validation and the test sets, respectively.
\item $\text{Valid}_{\text{full}}$, $\text{Test}_{\text{full}}$: the data consisting of the original validation and test sets, respectively.  
%$\text{Valid}_{\text{private}}$ $+$ $\text{Valid}_{ \text{public}}$,  $\text{Test}_{ \text{private}}$ $+$  $\text{Test}_{\text{public}}$.
\end{enumerate}
%Note that validation set for early stopping is also generated on the main server like the training set.

\textbf{Training the main server and user device LMs}: The main server is trained with the public data (or in-house data), which do not includes words for representing months, days of the week, and seasons. 
Before the training of the user device with the private data, we considered two different network parameter initialization. The first is to randomly initialize the user device LM weight parameters, and the second is kind of a transfer learning~\cite{pan2010survey}, where the weight parameters of the user device LM is copied from the main server LM, which was trained with public (or in-house) data. For the latter method, the size of the main server LM and that of the user device LMs must be the same size and structure, so the LM for all user devices and the main server adopt 256x2 LSTM RNN’s in this experiment. \tablename~\ref{table:3} shows the results of the user device LM trained using private data in both initial states and also the performance when trained using all data (private + public). The dataset that each user device LM uses during training is summarized as follows:
\begin{enumerate}[nolistsep]
\item $\text{T}_{\text{full}}$: train with the public and private data simultaneously
\item $\text{T}_{\text{transfer}}$: pretrain with the public data and then retrain (fine-tune) with the private data
\item $\text{T}_{\text{private}}$: train with the private data only
\end{enumerate}

\subsubsection{Evaluation of SDGKT with a Single User Device LM}

%\begin{table}[t]
%\centering
%\caption{BPC on the validation and the test set for user device LM.}
%\label{table:3}
%\begin{tabular}{ccccccc}
%\hline\hline
%        & \multicolumn{2}{c}{Full data} & \multicolumn{2}{c}{Load init} & \multicolumn{2}{c}{Rand init} \\ \hline
%        & Valid             & Test             & Valid          & Test          & Valid          & Test          \\ \hline
%Full    & 1.271             & 1.275            & 1.395          & 1.397         & 1.486          & 1.490         \\ \hline
%Private & 1.151             & 1.167            & 1.131          & 1.147         & 1.177          & 1.198         \\ \hline
%Public  & 1.285             & 1.288            & 1.426          & 1.427         & 1.522          & 1.524         \\ \hline\hline
%\end{tabular}
%\end{table}
\begin{table}[t]
\centering
\caption{BPC on the validation and the test set for a user device LM with the standard training.}
\label{table:3}
\begin{tabular}{ccccc}
\hline\hline
                           &       & Full  & Private & Public \\ \hline
\multirow{2}{*}{$\text{T}_{\text{full}}$} & Valid & 1.271 & 1.151   & 1.285  \\ 
                           & Test  & 1.275 & 1.167   & 1.288  \\ \hline
\multirow{2}{*}{$\text{T}_{\text{transfer}}$} & Valid & 1.395 & 1.131   & 1.426  \\ 
                           & Test  & 1.397 & 1.147   & 1.427  \\ \hline
\multirow{2}{*}{$\text{T}_{\text{private}}$} & Valid & 1.486 & 1.177   & 1.522  \\ 
                           & Test  & 1.490 & 1.198   & 1.524  \\ \hline\hline
\end{tabular}
\end{table}

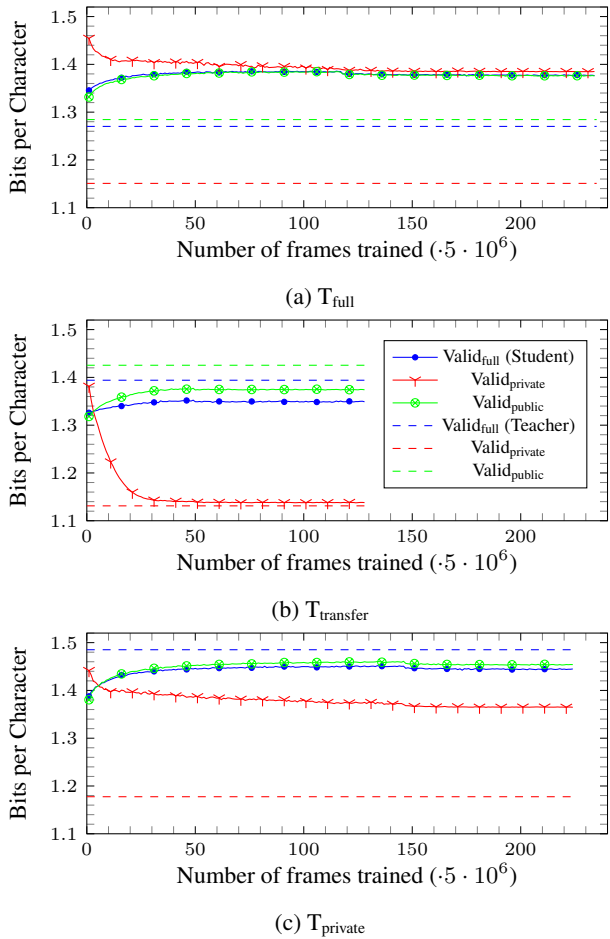
\begin{figure}[t]
\centering
\begin{subfigure}{.496\textwidth}
\begin{tikzpicture}
    \begin{axis}[
	width=\linewidth,
	height = 0.5\linewidth,
	%compat=1.12,
	xmin=0,
	ymin=1.1,
	xmax=240,
	ymax=1.52,
	label style={font=\footnotesize},
	legend style={font=\tiny,at={(0.8,0.5)},anchor=north west},
	tick label style={font=\scriptsize}, 
	domain=1:512, 
	minor x tick num=4, 
	minor y tick num=4, 
	log basis x={10}, 
	xtick pos=both, 
	xtick align=inside, 
	major tick style={line width=0.010cm, black},
	 major tick length=0.10cm,%same as default ]% 
        xlabel=Number of frames trained ($\cdot5\cdot10^6$),
	y label style={at={(axis description cs:0.05,0.5)}},
	x label style={at={(axis description cs:0.5,0.1)}},
        ylabel=Bits per Character]
	%\legend{Full data, Private data, Public data};
	\addplot[color=blue, mark=*, mark size=1.0pt, solid, mark repeat=15,mark options=solid] file{data_privacy/data_256x2_teacher_distill_full_devsplit_forcenewline_neweval_fulldatadev.txt}; 
	\addplot[color=red, mark=Mercedes star flipped, mark size=2.5pt, solid, mark repeat=10,mark options=solid] file{data_privacy/data_256x2_teacher_distill_full_devsplit_forcenewline_neweval_monthdatadev.txt}; 
	\addplot[color=green, mark=otimes, mark size=1.6pt, solid, mark repeat=15,mark options=solid] file{data_privacy/data_256x2_teacher_distill_full_devsplit_forcenewline_neweval_nomonthdatadev.txt};
	\addplot[color=blue, mark=none, dashed, mark repeat=15,mark options=solid] coordinates{(0,1.270148) (235,1.270148)};
	\addplot[color=red, mark=none, dashed, mark repeat=15,mark options=solid] coordinates{(0,1.150908) (235,1.150908)};
	\addplot[color=green, mark=none, dashed, mark repeat=15,mark options=solid] coordinates{(0,1.284453) (235,1.284453)};
    \end{axis}
   \end{tikzpicture}
 \caption{$\text{T}_{\text{full}}$}\label{fig_5_a}
\end{subfigure}

\begin{subfigure}{.496\textwidth}
\begin{tikzpicture}
    \begin{axis}[
	width=\linewidth,
	height = 0.5\linewidth,
	%compat=1.12,
	xmin=0,
	ymin=1.1,
	xmax=240,
	ymax=1.52,
	label style={font=\footnotesize},
	legend style={nodes={scale=0.7, transform shape},at={(0.57,0.9)},anchor=north west},
	tick label style={font=\scriptsize}, 
	domain=1:512, 
	minor x tick num=4, 
	minor y tick num=4, 
	log basis x={10}, 
	xtick pos=both, 
	xtick align=inside, 
	major tick style={line width=0.010cm, black},
	 major tick length=0.10cm,%same as default ]% 
        xlabel=Number of frames trained ($\cdot5\cdot10^6$),
	y label style={at={(axis description cs:0.05,0.5)}},
	x label style={at={(axis description cs:0.5,0.1)}},
        ylabel=Bits per Character]
	\legend{$\text{Valid}_{\text{full}}$ (Student), $\text{Valid}_{\text{private}}$, $\text{Valid}_{\text{public}}$, $\text{Valid}_{\text{full}}$ (Teacher), $\text{Valid}_{\text{private}}$, $\text{Valid}_{\text{public}}$};
	\addplot[color=blue, mark=*, mark size=1.0pt, solid, mark repeat=15,mark options=solid] file{data_privacy/data_256x2_teacher_distill_load_devsplit_forcenewline_neweval_fulldatadev.txt}; 
	\addplot[color=red, mark=Mercedes star flipped, mark size=2.5pt, solid, mark repeat=10,mark options=solid] file{data_privacy/data_256x2_teacher_distill_load_devsplit_forcenewline_neweval_monthdatadev.txt}; 
	\addplot[color=green, mark=otimes, mark size=1.6pt, solid, mark repeat=15,mark options=solid] file{data_privacy/data_256x2_teacher_distill_load_devsplit_forcenewline_neweval_nomonthdatadev.txt};
	\addplot[color=blue, mark=none, dashed, mark repeat=15,mark options=solid] coordinates{(0,1.394051) (128,1.394051)};
	\addplot[color=red, mark=none, dashed, mark repeat=15,mark options=solid] coordinates{(0,1.131159) (128,1.131159)};
	\addplot[color=green, mark=none, dashed, mark repeat=15,mark options=solid] coordinates{(0,1.425397) (128,1.425397)};
    \end{axis}
   \end{tikzpicture}
 \caption{$\text{T}_{\text{transfer}}$}\label{fig_5_b}
\end{subfigure}

\begin{subfigure}{.496\textwidth}
\begin{tikzpicture}
    \begin{axis}[
	width=\linewidth,
	height = 0.5\linewidth,
	%compat=1.12,
	xmin=0,
	ymin=1.1,
	xmax=240,
	ymax=1.52,
	label style={font=\footnotesize},
	legend style={font=\tiny,at={(0.8,0.5)},anchor=north west},
	tick label style={font=\scriptsize}, 
	domain=1:512, 
	minor x tick num=4, 
	minor y tick num=4, 
	log basis x={10}, 
	xtick pos=both, 
	xtick align=inside, 
	major tick style={line width=0.010cm, black},
	 major tick length=0.10cm,%same as default ]% 
        xlabel=Number of frames trained ($\cdot5\cdot10^6$),
        ylabel=Bits per Character,
	x label style={at={(axis description cs:0.5,0.1)}},
	y label style={at={(axis description cs:0.05,0.5)}}]
	%\legend{Full data,Private data, Public data};
	\addplot[color=blue, mark=*, mark size=1.0pt, solid, mark repeat=15,mark options=solid] file{data_privacy/data_256x2_teacher_distill_rand_devsplit_forcenewline_neweval_fulldatadev.txt}; 
	\addplot[color=red, mark=Mercedes star flipped, mark size=2.5pt, solid, mark repeat=10,mark options=solid] file{data_privacy/data_256x2_teacher_distill_rand_devsplit_forcenewline_neweval_monthdatadev.txt}; 
	\addplot[color=green, mark=otimes, mark size=1.6pt, solid, mark repeat=15,mark options=solid] file{data_privacy/data_256x2_teacher_distill_rand_devsplit_forcenewline_neweval_nomonthdatadev.txt};
	\addplot[color=blue, mark=none, dashed, mark repeat=15,mark options=solid] coordinates{(0,1.485148) (226,1.485148)};
	\addplot[color=red, mark=none, dashed, mark repeat=15,mark options=solid] coordinates{(0,1.177536) (226,1.177536)};
	\addplot[color=green, mark=none, dashed, mark repeat=15,mark options=solid] coordinates{(0,1.521800) (226,1.521800)};
    \end{axis}
   \end{tikzpicture}
 \caption{$\text{T}_{\text{private}}$}\label{fig_5_c}
\end{subfigure}

\caption{Convergence curves in terms of BPC on validation set for the single-cycle training. Dotted lines represent the BPC of the teacher networks which are reported in \tablename~\ref{table:3}, and solid lines show the BPC of the student networks with SDGKT training.} \label{fig_5}
\end{figure}

\begin{figure}[t]
\centering

\begin{subfigure}{.496\textwidth}
\begin{tikzpicture}
    \begin{axis}[
	width=\linewidth,
	height = 0.5\linewidth,
	%compat=1.12,
	xmin=0,
	ymin=1.1,
	xmax=240,
	ymax=1.52,
	label style={font=\footnotesize},
	legend style={nodes={scale=0.7, transform shape},at={(0.57,0.9)},anchor=north west},
	tick label style={font=\scriptsize}, 
	domain=1:512, 
	minor x tick num=4, 
	minor y tick num=4, 
	log basis x={10}, 
	xtick pos=both, 
	xtick align=inside, 
	major tick style={line width=0.010cm, black},
	 major tick length=0.10cm,%same as default ]% 
        xlabel=Number of frames trained ($\cdot5\cdot10^6$),
	x label style={at={(axis description cs:0.5,0.1)}},
	 y label style={at={(axis description cs:0.05,0.5)}},
        ylabel=Bits per Character]
	\legend{$\text{Valid}_{\text{full}}$ (Student), $\text{Valid}_{\text{private}}$, $\text{Valid}_{\text{public}}$, $\text{Valid}_{\text{full}}$ (Teacher), $\text{Valid}_{\text{private}}$, $\text{Valid}_{\text{public}}$};
	\addplot[color=blue, mark=*, mark size=1.0pt, solid, mark repeat=15,mark options=solid] file{data_privacy/data_256x2_teacher_full_infinite_fulldata.txt}; 
	\addplot[color=red, mark=Mercedes star flipped, mark size=2.5pt, solid, mark repeat=10,mark options=solid] file{data_privacy/data_256x2_teacher_full_infinite_monthdata.txt}; 
	\addplot[color=green, mark=otimes, mark size=1.6pt, solid, mark repeat=15,mark options=solid] file{data_privacy/data_256x2_teacher_full_infinite_nomonthdata.txt};
        \addplot[color=blue, mark=none, dashed, mark repeat=15,mark options=solid] coordinates{(0,1.270148) (62,1.270148)};
	\addplot[color=red, mark=none, dashed, mark repeat=15,mark options=solid] coordinates{(0,1.150908) (62,1.150908)};
	\addplot[color=green, mark=none, dashed, mark repeat=15,mark options=solid] coordinates{(0,1.284453) (62,1.284453)};
    \end{axis}
   \end{tikzpicture}
 \caption{$\text{T}_{\text{full}}$}\label{fig_6_a}
\end{subfigure}

\begin{subfigure}{.496\textwidth}
\begin{tikzpicture}
    \begin{axis}[
	width=\linewidth,
	height = 0.5\linewidth,
	%compat=1.12,
	xmin=0,
	ymin=1.1,
	xmax=240,
	ymax=1.53,
	label style={font=\footnotesize},
	legend style={font=\tiny,at={(0.5,0.5)},anchor=north west},
	tick label style={font=\scriptsize}, 
	domain=1:512, 
	minor x tick num=4, 
	minor y tick num=4, 
	log basis x={10}, 
	xtick pos=both, 
	xtick align=inside, 
	major tick style={line width=0.010cm, black},
	 major tick length=0.10cm,%same as default ]% 
        xlabel=Number of frames trained ($\cdot5\cdot10^6$),
        ylabel=Bits per Character,
	x label style={at={(axis description cs:0.5,0.1)}},
 y label style={at={(axis description cs:0.05,0.5)}}]
	%\legend{Full data, Private data, Public dataa};
	\addplot[color=blue, mark=*, mark size=1.0pt, solid, mark repeat=15,mark options=solid] file{data_privacy/data_256x2_teacher_loadinit_infinite_fulldata.txt}; 
	\addplot[color=red, mark=Mercedes star flipped, mark size=2.5pt, solid, mark repeat=10,mark options=solid] file{data_privacy/data_256x2_teacher_loadinit_infinite_monthdata.txt}; 
	\addplot[color=green, mark=otimes, mark size=1.6pt, solid, mark repeat=15,mark options=solid] file{data_privacy/data_256x2_teacher_loadinit_infinite_nomonthdata.txt};
	\addplot[color=blue, mark=none, dashed, mark repeat=15,mark options=solid] coordinates{(0,1.394051) (91,1.394051)};
	\addplot[color=red, mark=none, dashed, mark repeat=15,mark options=solid] coordinates{(0,1.131159) (91,1.131159)};
	\addplot[color=green, mark=none, dashed, mark repeat=15,mark options=solid] coordinates{(0,1.425397) (91,1.425397)};
    \end{axis}
   \end{tikzpicture}
 \caption{$\text{T}_{\text{transfer}}$}\label{fig_6_b}
\end{subfigure}

\begin{subfigure}{.496\textwidth}
\begin{tikzpicture}
    \begin{axis}[
	width=\linewidth,
	height = 0.5\linewidth,
	%compat=1.12,
	xmin=0,
	ymin=1.1,
	xmax=240,
	ymax=1.53,
	label style={font=\footnotesize},
	legend style={font=\tiny,at={(0.5,0.5)},anchor=north west},
	tick label style={font=\scriptsize}, 
	domain=1:512, 
	minor x tick num=4, 
	minor y tick num=4, 
	log basis x={10}, 
	xtick pos=both, 
	xtick align=inside, 
	major tick style={line width=0.010cm, black},
	major tick length=0.10cm,%same as default ]% 
        xlabel=Number of frames trained ($\cdot5\cdot10^6$),
	 y label style={at={(axis description cs:0.05,0.5)}},
	 x label style={at={(axis description cs:0.5,0.1)}},
        ylabel=Bits per Character]
%	xlabel shift=-3pt,
%	ylabel shift=-3pt]
	%\legend{Full data, Private data, Public data};
	\addplot[color=blue, mark=*, mark size=1.0pt, solid, mark repeat=15,mark options=solid] file{data_privacy/data_256x2_teacher_randinit_infinite_fulldata.txt}; 
	\addplot[color=red, mark=Mercedes star flipped, mark size=2.5pt, solid, mark repeat=10,mark options=solid] file{data_privacy/data_256x2_teacher_randinit_infinite_monthdata.txt}; 
	\addplot[color=green, mark=otimes, mark size=1.6pt, solid, mark repeat=15,mark options=solid] file{data_privacy/data_256x2_teacher_randinit_infinite_nomonthdata.txt};
	\addplot[color=blue, mark=none, dashed, mark repeat=15,mark options=solid] coordinates{(0,1.485148) (84,1.485148)};
	\addplot[color=red, mark=none, dashed, mark repeat=15,mark options=solid] coordinates{(0,1.177536) (84,1.177536)};
	\addplot[color=green, mark=none, dashed, mark repeat=15,mark options=solid] coordinates{(0,1.521800) (84,1.521800)};
    \end{axis}
   \end{tikzpicture}
 \caption{$\text{T}_{\text{private}}$}\label{fig_6_c}
\end{subfigure}
\caption{Convergence curves in terms of BPC on validation set for the multi-cycle training. Dotted lines represent the BPC of the teacher networks which are reported in \tablename~\ref{table:3}, and solid lines show the BPC of the student networks with SDGKT training.} \label{fig_6}
\end{figure}
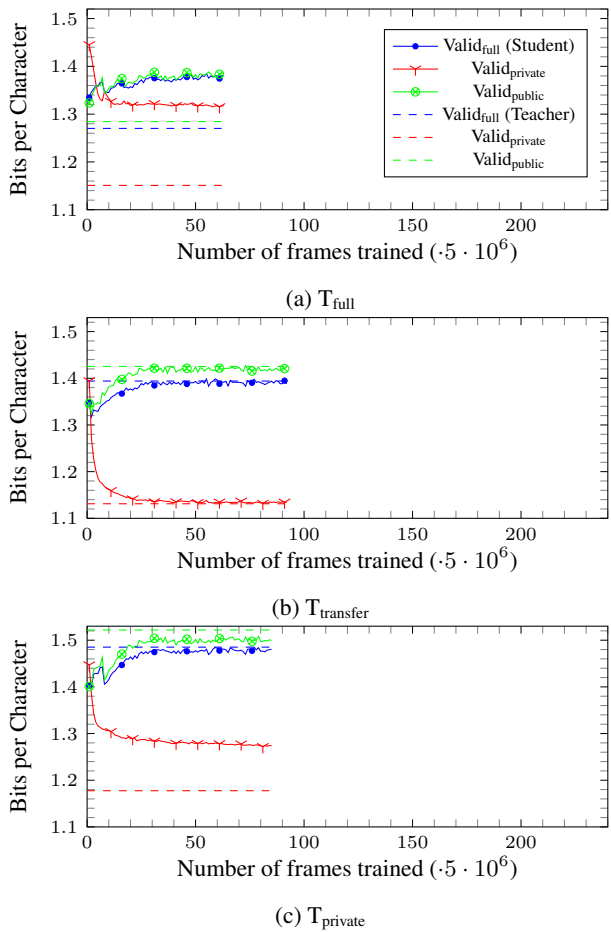

We first evaluate SDGKT in~\figurename~\ref{fig:1_b} with a small user device LM (teacher network)\footnote{This experiment is the simple example of SDGKT mentioned in Section~\ref{subsec:student}.}. Specifically, two experiments are conducted to examine the influence of the `cycle' mentioned in Section~\ref{subsec:student}. In the first experiment, the student network is trained with 200M characters of generated data with a single-cycle of SDGKT. On the other hand, in the second experiment, SDGKT is performed through multiple cycles, where 5M training characters are generated per cycle (`lot'). \figurename~\ref{fig_5} and \figurename~\ref{fig_6} show the convergence curve of SDGKT by employing three different types of user device LMs.

From the results of the single-cycle experiment in~\figurename~\ref{fig_5}, we observed that the OOV words can be learned only when `$\text{T}_{\text{transfer}}$' is employed as a teacher network in the user device. However, in the multi-cycle training results in~\figurename~\ref{fig_6}, the OOV words can be learned with all the three types of teacher networks, and the training speed is much faster compared to the single-cycle experiment. This is because the OOV words cannot appear in the generated training text with SDGKT in the initial cycle of the training. Therefore, OOV words such as `JANUARY' or `DECEMBER' are difficult to be learned with the single-cycle training. Even if the soft labels are generated from the teacher network, which is aware of the OOV words, the soft labels generated in the initial cycle only contains the information of single character prediction probabilities from the given generated text. However, in the multi-cycle experiments, the partial OOV words such as ``JANU, JANUA, and JUNUAR'' gradually appear in the generated text as the cycle progresses, and as a result, all the three student networks are able to learn OOV words.

Nevertheless, even in the single-cycle experiment the `$\text{T}_{\text{transfer}}$' was able to teach OOV words to the student network. Interestingly, the similar observation was reported by \citet{hinton2015distilling} with feedforward deep neural networks. In this experiment, knowledge transfer was performed between two digit classification networks, where it was shown that even if images of the number 3 was not given to the teacher network during the knowledge transfer, the student network was able to correctly classify the image of the number 3 to some extent. However, further analysis is required to reach the general explanation for this observation in the case of RNNs.

As the training progresses, the BPC measured on the private validation set gradually improves with the sacrifice of the BPC on the validation set of the public data. This is because the student network in the main server, which is trained with the public data, becomes adapted to the distribution of the private training data that is used for training the user device LM. As a result, the student network is fine-tuned to the user data without accessing this data directly. \tablename~\ref{table:4} shows the results on the three test sets with the three types of teacher networks. From the table, it is observed that the user device LM trained with `$\text{T}_{\text{transfer}}$' method shows the best performance on the private test data, which means that the transfer learning method is suitable for training the user device LM if the gradual adaptation of the main server LM to the private user data is the main concern. Therefore, we only employ `$\text{T}_{\text{transfer}}$' for training the user device LMs in the following multi user device experiments in Section~\ref{subsubsec:GKT}.
%\begin{table}[t]
%\centering
%\caption{BPC on the test set for 1 cycle and multiple cycles SGKT.}
%\label{table:4}
%\begin{tabular}{ccccccc}
%\hline\hline
%        & \multicolumn{2}{c}{Full data} & \multicolumn{2}{c}{Load init} & \multicolumn{2}{c}{Rand init} \\ \hline
%\# of cycle        & 1      & Multi     & 1      & Multi     & 1      & Multi     \\ \hline
%Full    & 1.356        & 1.376           & 1.352        & 1.390           & 1.447        & 1.479           \\ \hline
%Private & 1.396        & 1.330           & 1.153        & 1.148           & 1.381        & 1.286           \\ \hline
%Public  & 1.351        & 1.381           & 1.375        & 1.419           & 1.455        & 1.501           \\ \hline\hline
%\end{tabular}
%\end{table}
\begin{table}[t]
\centering
\caption{BPC on the test sets with the single-cycle and multi-cycle SDGKT methods using a single teacher network.}
\label{table:4}
\begin{tabular}{ccccc}
\hline\hline
                           & \# of cycles & $\text{Test}_{\text{full}}$ & $\text{Test}_{\text{private}}$ & $\text{Test}_{\text{public}}$ \\ \hline
\multirow{2}{*}{$\text{T}_{\text{full}}$} & Single    & 1.356 & 1.396   & 1.351  \\  
                           & Multi        & 1.376 & 1.330   & 1.381  \\ \hline
\multirow{2}{*}{$\text{T}_{\text{transfer}}$} & Single            & 1.352 & 1.153   & 1.375  \\  
                           & Multi        & 1.390 & 1.148   & 1.419  \\ \hline
\multirow{2}{*}{$\text{T}_{\text{private}}$} & Single     & 1.447 & 1.381   & 1.455  \\  
                           & Multi        & 1.479 & 1.286   & 1.501  \\ \hline\hline
\end{tabular}
\end{table}

\subsubsection{GKT with the Ensemble of User Device LMs}
\label{subsubsec:GKT}

In this section, experiments are conducted with multiple user LMs for language model adaptation while minimizing privacy infringement. For the experiments, we divide the private data into 100 sets, where each set is exclusively given for training a single user device LM. As a result, each user device LM is trained with only about 300K characters of private data. In the following experiments, we set the number of user devices to 10, 50, and 100. The soft labels, which are delivered to the main server LM, are obtained by averaging the output soft target values generated from the user device LMs by employing the trusted third party. In brief, the main server LM is trained with the generated text sequence as the input and the ensemble of the corresponding soft targets as the target. \tablename~\ref{table:5} shows the averaged BPC over the user device LMs and the ensemble BPC, which is obtained from the aggregated outputs from the user device LMs.
\tablename~\ref{table:6} shows the training result of the main server LM with the SDGKT and TDGKT schemes. Note that `T' indicates the TDGKT scheme, and `S' denotes the SDGKT scheme, which are described in~\figurename~\ref{fig:1}. By comparing the results with the ones in \tablename~\ref{table:5}, we can see that BPC of the main server LM is better than the average BPC of the user device LMs since the main server LM is trained with the ensemble of the user device LMs. As mentioned in Section~\ref{sec:application}, our proposed system employs SDGKT over TDGKT for stronger privacy protection, however, the result of the TDGKT training is also reported for comparison. In the case of the TDGKT, the training text is generated by a randomly selected user device and broadcasted to the other user devices. As can be seen in \tablename~\ref{table:6}, both the TDGKT and SDGKT method showed very similar performance.

The BPC of the LM utilizing the private data improves when increasing the number of devices. The performance with 50 devices was slightly better than that with 10 devices, whereas it is not much different from that with 100 devices. Compared with the performance of the server LM, which is pretrained using only public data, the BPC of the server LM on the private test data is improved after SDGKT. On the other hand, the BPC is slightly degraded on the public test data. This is because the server LM becomes adapted to the words and grammars in the private user data by slightly forgetting the pretrained expressions that is not frequently used in the user data. Note that as in the single user device experiments, the server LM can follow the language trends by gradually learning OOV words such as newly coined words or trendy expressions.

\begin{table}[t]
\centering
\caption{BPC on the test set for teacher networks. `A' is average of the teacher’s BPCs’, and `E' is BPC of the teachers’ ensemble.}
\label{table:5}
\begin{tabular}{ccccc}
\hline\hline
\# of Teachers       &   & $\text{Test}_{\text{full}}$ & $\text{Test}_{\text{private}}$ & $\text{Test}_{\text{public}}$ \\ \hline
\multirow{2}{*}{10}  & A & 1.345 & 1.269   & 1.358  \\ 
                     & E & 1.323 & 1.246   & 1.335  \\ \hline
\multirow{2}{*}{50}  & A & 1.347 & 1.269   & 1.359  \\  
                     & E & 1.321 & 1.242   & 1.333  \\ \hline
\multirow{2}{*}{100} & A & 1.346 & 1.269   & 1.359  \\ 
                     & E & 1.321      & 1.242   &1.333    \\ \hline\hline
\end{tabular}
\end{table}

\begin{table}[t]
\centering
\caption{BPC on the test set for main server LM. `T' and `S' indicate the TDGKT and the SDGKT. The BPC of the initial main server LM is 1.328, 1.496, and 1.301 for $\text{Test}_{\text{full}}$, $\text{Test}_{\text{private}}$, and $\text{Test}_{\text{public}}$ data respectively.}
\label{table:6}
\begin{tabular}{ccccc}
\hline\hline
\# of Teachers       &  & $\text{Test}_{\text{full}}$ & $\text{Test}_{\text{private}}$ & $\text{Test}_{\text{public}}$ \\ \hline
\multirow{2}{*}{10}  & T        & 1.325 & 1.252   & 1.336  \\
                     & S        & 1.324 & 1.250   & 1.336  \\ \hline
\multirow{2}{*}{50}  & T        & 1.323 & 1.247   & 1.335  \\ 
                     & S        & 1.323 & 1.248   & 1.334  \\ \hline
\multirow{2}{*}{100} & T        & 1.323 & 1.247   & 1.334  \\ 
                     & S        & 1.322 & 1.247   & 1.334  \\ \hline\hline
\end{tabular}
\end{table}

\section{Concluding Remarks}
\label{sec:concluding}
Throughout the paper, generative knowledge transfer (GKT) techniques are proposed for RNN LMs, where knowledge transfer from the teacher network to the student network is performed with the text data generated by one of the networks. In teacher-driven GKT (TDGKT), the training text is generated by the teacher network, whereas in student-driven GKT (SDGKT), the text generation is performed by the student network. Although the training text is generated by the student network in SDGKT, we showed that it is able to transfer the knowledge of the OOV words, which only the teacher network is aware of, to the student network by employing multi-cycle SDGKT. Also, the SDGKT provides strong privacy protection when applied to the presented privacy-preserving LM adaptation task between the main server and the multiple user devices. The experimental results show that SDGKT allows efficient transfer of the knowledge contained in the private user text data, such as newly coined words or trendy expressions, to the RNN LM in the main server without direct access to the user data, thereby preserving the privacy of the users.

\pagebreak



%\bibliography{icml2017}

\bibliographystyle{icml2017}

\end{document}